%% file: main.tex
\documentclass[conference, 10pt]{IEEEtran}
\IEEEoverridecommandlockouts
\usepackage{cite}
\usepackage{amsmath,amssymb,amsfonts}
\usepackage{multirow} 
\usepackage[ruled,lined]{algorithm2e}
\usepackage{graphicx}
\usepackage{hyperref}
\usepackage{booktabs} 
\graphicspath{{figs/}}
\usepackage{booktabs} %
\usepackage{textcomp}
\usepackage[table]{xcolor}
\usepackage{comment}
\def\BibTeX{{\rm B\kern-.05em{\sc i\kern-.025em b}\kern-.08em
    T\kern-.1667em\lower.7ex\hbox{E}\kern-.125emX}}
\newlength\savewidth
\definecolor{tabthird}{rgb}{1, 1, 0.7}
\definecolor{tabsecond}{rgb}{1, 0.85, 0.7}
\definecolor{tabfirst}{rgb}{1, 0.7, 0.7}

\input{macro}

\begin{document}

\title{\proj: Efficient-Coding--Inspired In-Sensor Compression for Edge Vision}

\input{authors}

\maketitle

\input{abstract}

\input{intro}

\input{bg}

\input{encoding}

\input{codesign}

\input{sensor}

\input{eval}

\input{related}

\input{conclusion}

\section*{Acknowledgment}
The work is partially supported by NSF Award \#2416375.
\bibliographystyle{ieeetr}
\bibliography{references}

\end{document}

%% file: macro.tex
\ifx\figurename\undefined \def\figurename{Figure}\fi
\renewcommand{\figurename}{Fig.}
\renewcommand{\paragraph}[1]{\textbf{#1} }

\newcommand{\Sect}[1]{Sec.~\ref{#1}}
\newcommand{\Fig}[1]{Fig.~\ref{#1}}
\newcommand{\Tbl}[1]{Tbl.~\ref{#1}}

\newcommand{\Eqn}[1]{Eqn.~\ref{#1}}

\newcommand{\mode}[1]{\underline{\textsc{#1}}\xspace}

\newcommand{\proj}{\textsc{SnapPix}\xspace}

\newcommand{\RNum}[1]{\uppercase\expandafter{\romannumeral #1\relax}}

%% file: authors.tex
\author{
\IEEEauthorblockN{Weikai Lin\textsuperscript{1,*}, Tianrui Ma\textsuperscript{2,3,*}, Adith Boloor\textsuperscript{3}, Yu Feng\textsuperscript{4,$\#$}, Ruofan Xing\textsuperscript{5}, Xuan Zhang\textsuperscript{3,5}, Yuhao Zhu\textsuperscript{1}}
\IEEEauthorblockA{\textsuperscript{1}University of Rochester \quad
\textsuperscript{2}Institute of Computing Technology, CAS\quad
\textsuperscript{3}Washington University in St. Louis \quad \\
\textsuperscript{4}Shanghai Jiao Tong University \quad
\textsuperscript{5}Northeastern University\\ 
wlin33@ur.rochester.edu, matianrui@ict.ac.cn, adith@wustl.edu, 
y-feng@sjtu.edu.cn, \\
\{xing.ruo, xuan.zhang\}@northeastern.edu, yzhu@rochester.edu}
\vspace{1mm}
\thanks{*Both authors contributed equally to this research.
\par
$^{\#}$Work done while at University of Rochester.}
}

%% file: abstract.tex
\begin{abstract}

Energy-efficient image acquisition on the edge is crucial for enabling remote sensing applications where the sensor node has weak compute capabilities and must transmit data to a remote server/cloud for processing.
To reduce the edge energy consumption, this paper proposes a sensor-algorithm co-designed system called \proj, which compresses raw pixels in the analog domain inside the sensor.
We use coded exposure (CE) as the in-sensor compression strategy as it offers the flexibility to sample, i.e., selectively expose pixels, both spatially and temporally.
\proj has three contributions.
First, we propose a task-agnostic strategy to learn the sampling/exposure pattern based on the classic theory of efficient coding.
Second, we co-design the downstream vision model with the exposure pattern to address the pixel-level non-uniformity unique to CE-compressed images.
Finally, we propose lightweight augmentations to the image sensor hardware to support our in-sensor CE compression.
Evaluating on action recognition and video reconstruction, 
\proj outperforms state-of-the-art video-based methods at the same speed while reducing the energy by up to 15.4$\times$.
We have open-sourced the code at: \href{https://github.com/horizon-research/SnapPix}{https://github.com/horizon-research/SnapPix}.

\end{abstract}

%% file: intro.tex
\section{Introduction}

Energy-efficient imaging is essential for sensing scenarios where the sensor node has weak compute capability and must transmit data to a nearby edge server or even cloud for processing.
This includes scenarios such as traffic monitoring~\cite{klotz2024minimalist}, remote satellite sensing~\cite{gadre2024adapting}, or even smartphone/extended reality devices that offload computations~\cite{xreal_air_2}.
In all these scenarios, the sensing node energy is dominated by two components: 1) the image sensor itself, which is in turn dominated by the read-out circuitry~\cite{feng2024blisscam, ma2023leca}
, and 2) the wireless data transmission from the sensing node, whose energy is known to outweigh computation and increases with transmission distance~\cite{desai2022camaroptera}. 

To address this energy bottleneck, recent research has explored in-sensor compression, where raw pixels are aggressively sampled before being digitized and transmitted out of the sensor~\cite{ma2023leca, feng2024blisscam, yoshida2019high, zhang2023improving, nair20243d}.
While promising, in-sensor compression faces two challenges.
\begin{itemize}
    \item 
First, it must address a fundamental dilemma, where the sensor output is both an \textit{energy} bottleneck and an \textit{information} bottleneck, which are directly in contention with each other.
Reducing energy consumption requires aggressive in-sensor compression, but over-compression can lead to significant losses in task accuracy.
    \item
Second, the compressed data must support diverse downstream tasks. For instance, a smart-city camera might need to process the same video stream for both action recognition and traffic monitoring. Similarly, in augmented reality (AR) applications, a single camera stream must handle tasks such as hand recognition and scene reconstruction~\cite{kwon2023xrbench}.
\end{itemize}

\begin{figure}[t]
    \centering
    \includegraphics[width=\columnwidth]{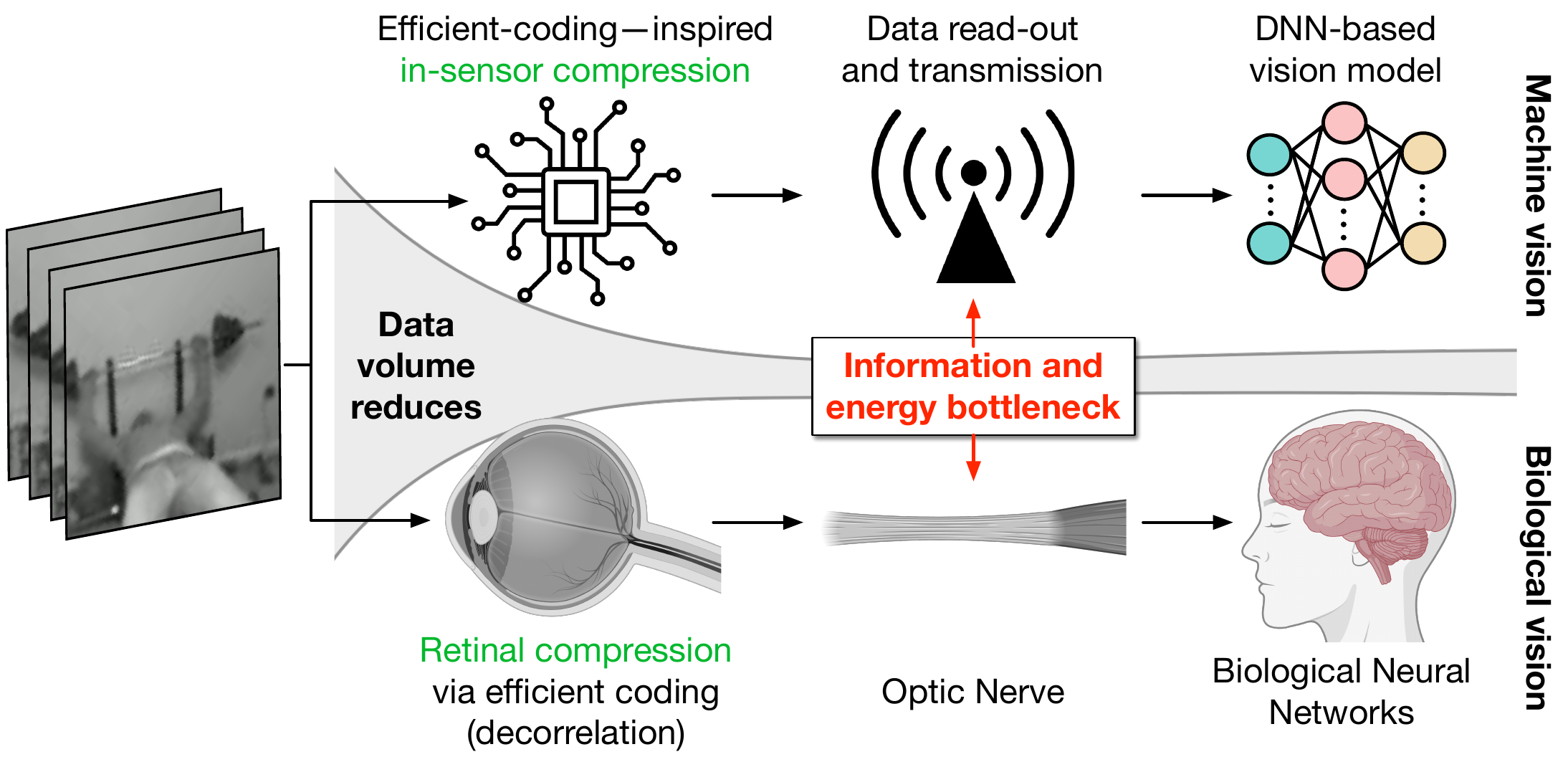}
    \caption{
    \proj reduces edge sensing energy through in-sensor compression by decorrelating output pixel values. This is inspired by the mammalian visual system, where the retina compresses information by decorrelating the retinal output neurons; signals carried through the optic nerve, while at a much lower bandwidth than at the initial stage of the retina, encodes essential information that permits the downstream visual cortex to effectively perform visual tasks.
    }
    \label{fig:cmv}
    \vspace{-10pt}
\end{figure}

The paper proposes \proj, a general in-sensor compression architecture that significantly reduces end-to-end system energy while maintaining high accuracy.
\proj achieves this through three key innovations, summarized below.

\paragraph{General-Purpose Sampling via Decorrelation.}
To design a general compression strategy, \proj leverages the principle of decorrelation to minimize redundancy among output pixels~(\Sect{sec:enc}).
This approach is inspired by the classic theory of efficient coding in neuroscience~\cite{barlow1961possible, attneave1954some, pitkow2012decorrelation, zbontar2021barlow}, which suggests that the retina efficiently transmits information to the brain by reducing redundancy and increasing decorrelation among the Retina Ganglion Cells (RGCs) --- the output neurons of the retina~\cite{dan1996efficient}.
This concept is visualized in \Fig{fig:cmv}.

Similarly, the compression pattern in \proj is trained to maximize information density in the sensor output, as opposed to be tailored to a particular task.
The core mechanism for compression in \proj is \textit{sampling}.
In particular, \proj samples through \textit{coded exposure} (CE)~\cite{reddy2011p2c2}, where the sensor selectively exposes pixels both spatially and temporally; pixel values are integrated across exposure slots into a single coded image before being read out~\cite{yoshida2019high}.

\paragraph{Co-Designed Vision Model and Coded Pattern.} 
To maximize the benefits of CE, ideally there should not be any constraint imposed on each pixel's exposure pattern.
That, however, means that pixels would carry varying amounts of information and should be treated differently in the downstream model, introducing performance overhead~\cite{ar_iccp, ar_tpami}.

Instead, \proj constrains the sampling pattern to be tile-repetitive --- pixels within a tile can have different exposure patterns, but the pattern repeats across tiles --- and thus constrains the pixel variation within a tile (\Sect{sec:codesign}).
To accommodate this tile-repetitive structure, we use Vision Transformers (ViTs)~\cite{dosovitskiy2021an} as the backbone. ViTs naturally process inputs tile-by-tile, and the processing of each tile is trained based on the (offline obtained) within-tile pixel variations.
With a carefully designed ViT architecture and tailored pre-training, our constrained sampling does not degrade accuracy.

\paragraph{Hardware Support.}
To support our in-sensor sampling strategy, we propose a set of minimal hardware augmentations to the stacked sensor architecture that is common in modern image sensors~\cite{oike2021evolution}, where the first layer is the pixel array and the second layer implements per-pixel digital logic (\Sect{sec:hw}).
The augmentations to each pixel are limited to a single-bit storage and two transistors to save and apply the CE pattern.
These augmentations are integrated beneath the pixel array, resulting in negligible area overhead.

\paragraph{Results.}
On various edge sensing scenarios (that differ in edge compute capabilities and in data transmission technologies), \proj achieves 1.4$\times$ to 15.4$\times$ energy saving compared to existing systems.
Compared to other compression methods with a similar compression rate, \proj delivers better accuracy on multiple tasks.

%% file: bg.tex
\section{Background}
\label{sec:bg}

\subsection{Energy Bottleneck in Edge Sensing}
\label{sec:bg:cmv}

This paper examines edge sensing scenarios where the edge comprises an image sensor and a lightweight compute unit, such as a microcontroller. In these setups, data must be offloaded to a server for computation-intensive processing. This paradigm is common in remote sensing applications, including urban sensing~\cite{adkins2018signpost} and satellite imaging~\cite{gadre2024adapting}.

The main energy bottleneck in such systems comes from the cost of in-sensor data read-out and sensor-host data transmission.
The read-out energy cost is dominated by the ADC. A recent survey shows that the ADC costs about 66\% of an image sensor's energy on average~\cite{ma2023leca}.
The data transmission includes both the MIPI CSI-2 interface, which transfers data from the sensor to the edge processor, and the subsequent transmission of data from the processor to the cloud. Energy-wise, transmitting one byte via the MIPI CSI-2 interface is approximately 300 times more expensive than performing a one-byte MAC operation~\cite{ma2023camj}. Wireless transmission exacerbates energy consumption, adding an order of magnitude to the cost, even with passive transmission methods like backscatter~\cite{talla2017lora}.

\subsection{Coded Exposure (CE)}
\label{sec:bg:ce}

CE compresses image data by selectively exposing pixels across spatial dimensions and multiple frames, integrating the exposures into a single coded image before readout~\cite{yoshida2019high}. 
\Fig{fig:ce} illustrates the fundamental operation of CE. Let \( Y \) represent the sequence of images that would have been captured using a conventional image sensor, with dimensions \( T \times H \times W \), where $T$ is the number of frames in $Y$, and $H$ and $W$ are the height and width of each image, respectively.

In the CE process, an exposure mask is applied to each frame (or exposure slot) \( t \) to select a subset of pixels to expose during that frame (e.g., 5 exposure slots as shown in \Fig{fig:ce}). 
The exposed pixel values are then integrated over time for each pixel, resulting in a single coded image of dimensions \( 1 \times H \times W \) being read out from the sensor. This approach achieves a data reduction factor of \( T \), as \( T \) frames are compressed into one.
This process can be formulated as follows:
\begin{equation}
\label{eqn:coded}
X(i, j) = \sum_{t=1}^{T} M(i, j, t) \cdot Y(i, j, t)
\end{equation}
\noindent where \( X \) is the final coded image,
\( M \) is the binary masks controlling exposure, \( i \), \( j \), and \( t \) index the spatial dimensions and the temporal dimension (frames), respectively.

%% file: encoding.tex
\section{General-Purpose Sampling via Decorrelation}
\label{sec:enc}

\begin{figure}[t]
    \centering
    \includegraphics[width=\columnwidth]{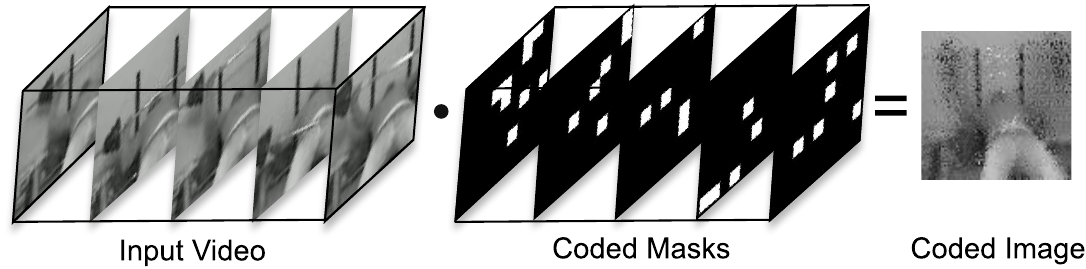}
    \caption{Coded exposure with 5 exposure slots. In each slot, pixels are selectively exposed, controlled by a coded mask. In the end, the values at all the exposure slots are integrated pixel-wise to form one single coded image.}
    \label{fig:ce}
    \vspace{-10pt}
\end{figure}

\begin{figure*}[t]
\centering
\includegraphics[width=0.77\textwidth]{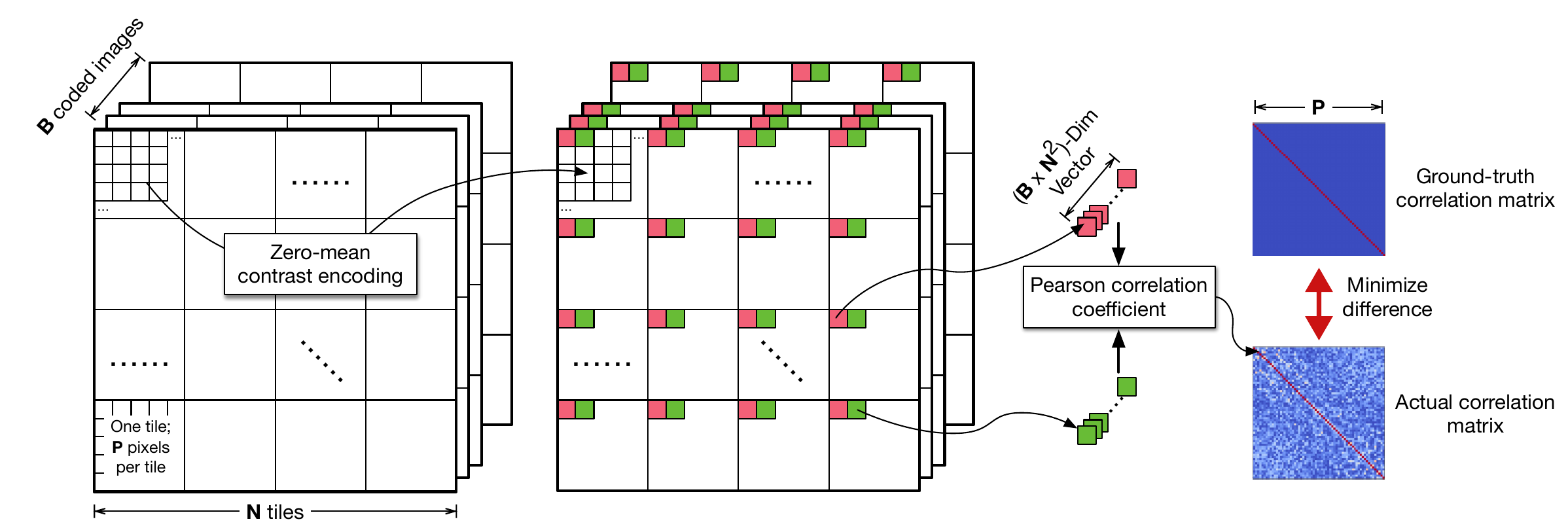}
\vspace{-2pt}
\caption{Illustration of training for pixel decorrelation in coded images. A coded image is divided into tiles, each containing $P$ pixels.
The CE mask (not shown) is optimized to decorrelate any pair of pixels within a tile. Zero-mean contrast encoding is applied, ensuring the mean pixel value of each tile is zero.
}
\label{fig:decor}
\vspace{-10pt}
\end{figure*}

Our CE pattern design maximizes decorrelation among pixels in coded images. Proximal pixels often exhibit high similarity and strong correlation, while pixels farther apart tend to be less correlated.
To focus on reducing redundancy among highly correlated pixels, we decorrelate pixels within a tile of \(P\) pixels.
This is achieved by minimizing the following loss:

\begin{equation}
\label{eq:decor_loss}
\mathcal{L}_{\text{Cor}} \triangleq \frac{1}{P(P - 1)} \sum_{i=1}^{P} \sum_{\substack{j \neq i}} C_{ij}^2
\end{equation}

\noindent where \( C_{ij} \) is the Pearson correlation coefficient between two distinct coded pixels \( i \) and \( j \) within a tile, quantifying redundancy between them. 

Estimating \( C_{ij} \) requires multiple samples for each coded pixel, and \Fig{fig:decor} illustrates this process.
First, we sample a batch of images from the dataset and apply the CE operation (\Eqn{eqn:coded}) to generate a batch of coded images.
Then we divide each coded image into \( N \times N \) tiles, yielding \( S = B \times N^2 \) samples per coded pixel, where \( B \) is the number of coded images in one batch and \( N^2 \) is the number of tiles per coded image.
As a result, each coded pixel in the batch is represented by an $S$-dimensional vector, which is used to calculate $C_{ij}$ for any pair of coded pixels $i$ and $j$.

Before forming the $S$-dimensional vectors, we preprocess each raw coded tile to have a zero mean, a step referred to as ``zero-mean contrast encoding'' in \Fig{fig:decor}.
Mathematically, for each tile in a raw coded image, we subtract the average pixel value of the tile from every pixel in the tile. 
The average value is computed by averaging across all the corresponding tiles in a dataset.
This zero-mean preprocessing is important, as proximal pixels naturally exhibit similar values and are thus highly correlated.
Without accounting for this inherent correlation, the training process can collapse, where all exposure slots remain closed for all pixels.

We learn the optimal exposure mask ($M$ in \Eqn{eqn:coded}) by minimizing $\mathcal{L}_{Cor}$.
This training is task-agnostic: the exposure mask is optimized for a given dataset irrespective of the downstream task.
We use straight-through estimation~\cite{bengio2013estimating} to propagate gradients through the binary masking operation.

%% file: codesign.tex
\section{CE-Optimized Vision Model and Training}
\label{sec:codesign}

\paragraph{Motivation.}
Pixels in coded images carry varying amounts of information due to their differences in exposure.
A pixel exposed in more slots retains richer temporal information and carries more motion blur.
Downstream models that ignore these variations experience accuracy degradation.

In particular, standard convolution operations treat all pixels uniformly, applying the same kernel to every pixel regardless of its exposure. 
Prior work~\cite{ar_iccp} proposed Shift-Variant Convolution (SVC) layer to address this issue by using different convolution kernels for different coded pixels. However, our profiling shows that SVC slows inference by 4$\times$ due to a lack of optimized implementations and inefficient GPU utilization.
Due to this performance issue, SVC has only been applied to the first layer in previous work~\cite{ar_iccp, ar_tpami}, negatively impacting the overall accuracy.

\begin{figure*}[t]
\centering
\includegraphics[width=0.77\textwidth]{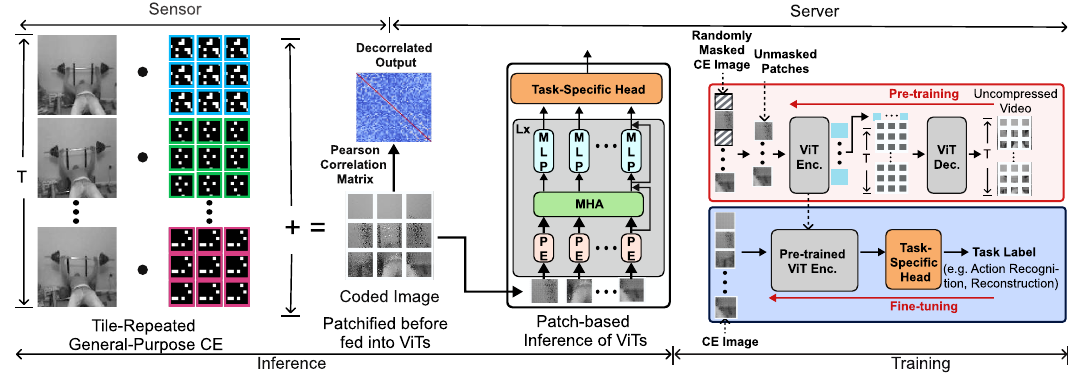}
\caption{The end-to-end pipeline of \proj with in-sensor CE for compression and a ViT-based vision model for downstream tasks.
The CE pattern is trained task-independently using decorrelation, while the downstream model is co-designed with CE patterns and pre-trained specifically for CE-encoded inputs.}
\label{fig:overview}
\vspace{-5pt}
\end{figure*}

\paragraph{CE-Optimized Vision Transformer.} 
If the exposure pattern is allowed to vary across all the pixels in a frame, then all the pixels are necessarily different.
Instead, we propose a tile-repetitive exposure pattern: the frame is divided into tiles, with pixel exposure variations allowed within each tile but constrained to be identical across all tiles. 
As shown in the left part of \Fig{fig:overview}: the exposure pattern is consistent across tiles for each time slot (three slots are illustrated), but it can change across slots.
As a result, the downstream models only have to deal with pixel variations within a tile while maintaining performance comparable to unconstrained patterns~\cite{ar_tpami}.

Building on this tile-repetitive pattern, we propose using Vision Transformers (ViTs)~\cite{dosovitskiy2021an} as the backbone for the downstream vision model.
ViTs naturally divide an image into patches of \(M \times M\) pixels and process pixels within each patch differently via patch-wise embedding (PE) and multi-layer perceptrons (MLPs), while enabling information sharing across patches through multi-head attention (MHA).
This makes ViTs well-suited for handling localized pixel variations, as illustrated in the middle part of \Fig{fig:overview}.

We set the CE tile size to match the ViTs’ patch size. This ensures MLPs can learn to address pixel-wise variations within each tile, which are determined by the decorrelation-based exposure pattern. To further facilitate ViT training, each pixel value is normalized by the number of exposure slots.

\paragraph{CE-Optimized Reconstruction Pre-training.} Inspired by recent reconstruction-based pre-training~\cite{he2022masked, wang2023videomae}, we propose a tailored pre-training procedure for CE-compressed inputs.

As illustrated in the right part of \Fig{fig:overview}, the pre-training process begins with a CE-coded image \( X \), sampled and integrated from a sequence of images. We randomly mask a large portion (e.g., 85\%) of its tiles, and set the objective of pre-training to reconstruct the original video sequence.
Unlike previous image-to-image~\cite{he2022masked} or video-to-video~\cite{wang2023videomae} 
pre-training approaches, our pre-training is designed to perform a ``coded image-to-video'' prediction. This requires the model to both predict the masked tiles, capturing spatial scene structure, and upsample temporal signals from CE-coded information, learning temporal dynamics.

The pre-training process can be formulated as follows:
\begin{equation}
    \hat{Y} = \text{D}(\text{E}(\text{random\_masking}(f(Y))))
\end{equation}
where \( Y \) is the original video, $f(\cdot)$ is CE function as defined in \Eqn{eqn:coded}, $E(\cdot)$ and $D(\cdot)$ are the pre-trained ViT encoder and decoder, and \( \hat{Y} \) is the reconstructed video.
We use the Mean Squared Error (MSE) loss for pre-training, and predict only 50\% of the video frames to accelerate pre-training~\cite{wang2023videomae}.

%% file: sensor.tex
\section{Hardware Support for In-Sensor CE}
\label{sec:hw}

\paragraph{Rationale and Overview.}
To support the CE function, previous solutions heavily augment each pixel, placing multiple floating diffusion (FD) nodes~\cite{gulve202339,luo202230}, using an arithmetic logic unit with mixed-signal memory~\cite{martel2020neural}, or communicating high-resolution CE pattern with off-chip controller~\cite{kagawa2022dual}.
They result in a larger pixel area and a smaller pixel fill factor.

In contrast, our sensor utilizes the characteristic of the tile-repetitive decorrelation pattern to reduce both the control footprint and control power.
In our design, we replace the global control with local, per-pixel storage to store the CE pattern.
We then use a die stacking design, where the top layer is the pixel array and the bottom layer hosts the digital logic/storage.
Such a stacking design is common in consumer CMOS image sensors~\cite{oike2021evolution}, and our design represents a novel usage of the bottom layer to support CE.
By stacking, the per-pixel storage is completely hidden beneath the pixel's exposure circuits, so the area of our proposed pixel is approximately the same as that of the conventional, non-CE pixel.

\begin{figure}[t]
    \centering
    \includegraphics[width=\columnwidth]{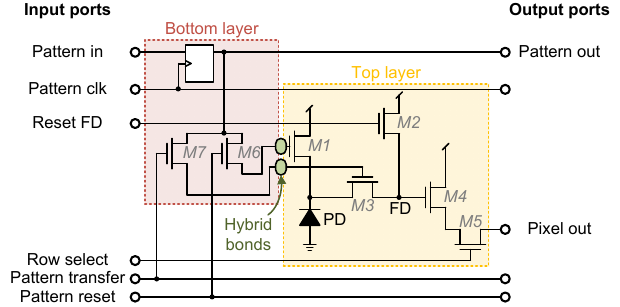}
    \caption{
    Schematic of the proposed CE pixel.
    It is based on a stacked design that is commonly used in modern CMOS image sensors~\cite{oike2021evolution}.
    }
    \label{fig:cepixel}
\vspace{-15pt}
\end{figure}

\paragraph{Design.}
The proposed pixel design is illustrated in \Fig{fig:cepixel}, which contains two layers.
The top layer is pixel array layer, where each pixel is based on the classic 4T Active Pixel Sensor (APS) design that consists of a photodiode (PD) to acquire incident light and five transistors to reset PD (\texttt{M1}), reset floating diffusion (FD) node (\texttt{M2}), transfer photocharge (\texttt{M3}), and read out photocharge as voltage when the pixel is selected (\texttt{M4}, \texttt{M5}).
Different to the conventional 4T pixel design, the additional transistor \texttt{M1} decouples the reset of PD and FD such that the PD can be exposed across multiple exposure slots but only transfer the photocharge once to the FD, thereby realizing CE function.

At the bottom layer, each pixel is equipped with a D-Flip-Flop (DFF) to buffer the one-bit CE pattern of a given exposure slot.
The bit represents whether to accumulate the pixel's exposure during the corresponding exposure slot.
The DFFs of all the pixels in a tile are connected in a shift-register style: the \textit{pattern in} wire of the second pixel is connected to the \textit{pattern out} wire of its preceding pixel.
Each pixel is also augmented with two additional transistors: \texttt{M6} (pattern reset) and \texttt{M7} (pattern transfer).

At the start of every exposure slot, the CE bits of each pixel in a tile are streamed in through the \textit{pattern in} wire and buffered in the corresponding DFFs .
Then, \texttt{M6} is turned on for all the pixels (through the \textit{pattern reset} wire), which allows the CE bit in the DFF to control the reset of the PD.
If the CE bit is 1, the PD is reset via \texttt{M1} (charges accumulated so far is cleared), getting ready for exposure; if the CE bit is 0, the PD is not reset (since \texttt{M1} will be open).
The DFFs can then be powered-gated.

After the exposure, the same CE bits are streamed in and buffered again in the DFFs.
We then turn on \texttt{M7} for all the pixels (through the \textit{pattern transfer} wire).
If the CE bit is 1, the charge is transferred from the PD to the FD through \texttt{M3}; otherwise, \texttt{M3} is open and the FD does not accumulate the charge from the previous exposure.
The DFFs are then power-gated again until the next exposure slot.
While the DFFs are power-gated, logic 0 is given to both \texttt{M1} and \texttt{M3} via a simple reset logic (not shown in \Fig{fig:cepixel} for simplicity).
The delay between the \textit{pattern reset} and \textit{pattern transfer} signals physically creates the exposure time for the pixel~\cite{yoshida2019high}.

\paragraph{Area Overhead.}
We synthesize the digital logic at the bottom layer in TSMC $65~nm$ and obtain the area estimation of $30~\mu m^2$.
According to DeepScale tool~\cite{sarangi2021deepscaletool}, it translates to $3.2~\mu m^2$ in $22~nm$, which is much smaller than commercial stacked digital pixel sensors (DPS)~\cite{ikeno20224, seo20222}. Therefore, the pixel area is constrained by the APS at the top layer rather than the introduced logic.

An alternative design is to broadcast the CE pattern to all the pixels in a tiles via dedicated wires.
This approach would remove the need for per-pixel digital logic, but require $2N$ signal wires per pixel given a tile size of $N\times N$. 
In contrast, the number of wires in our design is constant: only four (\textit{pattern in}, \textit{pattern clk}, \textit{pattern transfer}, and \textit{pattern reset}), regardless of tile size.
Our synthesis results show that when $N=8$, the signal wires occupy $2.24~\mu m \times 2.24~\mu m$; as $N$ increases to 14, the wire area grows to $3.92~\mu m \times 3.92~\mu m$, exceeding the area of the state-of-the-art APS.

%% file: eval.tex
\section{Evaluation}
\label{sec:eval}

\subsection{Experimental Setup} 
\label{sec:eval:exp}

\paragraph{Dataset.} 
\label{eval:dataset}
We evaluate our approach on three datasets: Something-Something v2 (SSV2)~\cite{goyal2017something}, Kinetics-400 (K400)~\cite{kay2017kinetics}, and UCF-101~\cite{soomro2012ucf101}, using
the first train-test split in UCF-101.
For pre-training, we use SSV2 and the larger K710 dataset~\cite{wang2023videomae}.
We downsample each video's shorter dimension to 112 pixels and convert the videos to grayscale in linear space.
We simulate a CE dataset using $T=16$ (recall $T$ is the number of exposure slots in \Eqn{eqn:coded}).
Each frame is center-cropped to produce a $112 \times 112$ resolution input, which is also the input resolution to \proj.
The video baselines accepts 16 consecutive, uncoded frames as their inputs (i.e., $16 \times 112 \times 112$ in resolution).

\paragraph{Tasks.} We evaluate two distinct downstream tasks: action recognition (AR), a high-level task producing a single classification output, and reconstruction (REC), a low-level task generating a video. 
REC evaluation addresses scenarios where videos are stored for future, undefined tasks. AR is tested on SSV2, K400, and UCF101 using clip-1 crop-1 accuracy, while REC is evaluated on SSV2, which reconstructs the original \(16 \times 112 \times 112\) frames from a \(112 \times 112\) coded image, with Peak Signal-to-Noise Ratio (PSNR) as the metric.

\paragraph{Variants.}
We provide two \proj variants: \proj-B and \proj-S, differing only in the backbone.
\proj-B uses ViT-B ($87$M parameters), for higher accuracy
but at a slower speed. \proj-S uses ViT-S ($22$M parameters) for faster performance, similar to prior CE-based models~\cite{ar_iccp, ar_tpami}. Both variants use an \(8\times8\) patch size.

\paragraph{Baselines.}
We compare our decorrelated CE pattern with the following task-agnostic CE patterns ($T=16$ in all cases):
\begin{itemize}
\setlength{\itemindent}{-1em}
    \item \mode{Long Exposure}: All pixels exposed in all slots.
    \item \mode{Short Exposure}: All pixels exposed every 8th frame.
    \item \mode{Random}: Each pixel exposed randomly with a 50\% probability per exposure slot.
    \item \mode{Sparse Random}: Each pixel exposed once randomly across $T$ exposure slots.
\end{itemize}

For AR, we also compare against three prior systems:

\begin{itemize}
\setlength{\itemindent}{-1em}
    \item \mode{SVC2D}~\cite{ar_iccp}: A CE-based AR model with SVC and an end-to-end learned CE pattern.
    \item \mode{C3D}~\cite{tran2015learning}: A strong video-based AR model, treated as an upper bound in prior CE works~\cite{ar_iccp, ar_tpami}.
    \item \mode{VideoMAEv2-ST}~\cite{wang2023videomae}: A state-of-the-art ViT-based video AR model, adjusted to match \proj-B's speed.
\end{itemize}

To ensure a fair comparison, we reproduced all baselines using our data preprocessing, frame sampling strategy, and training recipe, modifying only the learning rate.

\begin{figure}[t]
    \centering
    \includegraphics[width=0.49\textwidth]{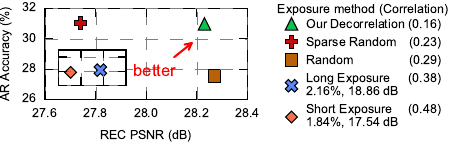}
    \caption{Comparison of task-agnostic CE patterns using AR and REC results. The legend also shows the corresponding Pearson correlation coefficients.}
    \label{fig:ce_comp}
    \vspace{-5pt}
\end{figure}

\paragraph{Training Recipe.}
To avoid overfitting to smaller datasets, we first train the decorrelated CE pattern on the large pre-training dataset for 5 epochs. 
We then fix the CE pattern and perform pre-training and task-specific training  for downstream vision models.
For training from scratch, REC is trained for 100 epochs, AR for 200 epochs on SSV2, and 400 epochs on K400 and UCF-101. For fine-tuning after pre-training, epochs are halved. 
Epochs are computed as repeated augmentations$\times$epochs, same as prior work~\cite{wang2023videomae}. Learning rates are tuned for different models. %

\subsection{Decorrelation Outperforms Other Task-Agnostic CEs}
\label{sec:eval:decor}

To evaluate the effectiveness of our decorrelation-based CE (\Sect{sec:enc}), we compare it against other task-agnostic CE patterns in \Fig{fig:ce_comp}, where reports the AR accuracy ($y$-axis) and the REC quality ($x$-axis) on the SSV2 dataset using our CE-optimized ViT (\Sect{sec:codesign}) trained from scratch.
The legend on the right also reports the Pearson correlation coefficient between coded pixels.

Our decorrelation-based CE excels in both AR and REC, demonstrating its effectiveness as a general task-agnostic CE pattern.
While \mode{Random} performs slightly better in REC, its AR accuracy is significantly lower than ours.
Conversely, \mode{Sparse Random} achieves a high AR accuracy but performs poorly in REC.
\mode{Long Exposure} and \mode{Short Exposure} perform significantly worse in both tasks.
The relative performance across these strategies corresponds roughly to their respective correlation coefficient, suggesting the benefit of using decorrelation to learn the CE pattern.

\subsection{\proj Outperforms Task-Specific Methods}
\label{sec:eval:acc}

We now evaluate \proj against prior systems that are specifically tuned for AR, including CE-based and video-based methods.
Inference speed is measured in inference per second on a system with a Ryzen 9 7900X3D CPU and a RTX 4090 GPU using a batch size of 64, simulating server-side computations.
Results are summarized in \Tbl{tab:model_comparison}. 
\mode{\proj-B} and \mode{\proj-S} achieve the highest AR accuracy across all datasets, nearly doubling the performance of CE-based baselines. They also surpass video models, which were previously considered upper bounds for CE-based methods~\cite{ar_iccp, ar_tpami}.
At the same time, \proj executes faster than video-based methods due to a reduced input data.

It might initially seem surprising that \proj out-performs even the video baselines (last two rows in \Tbl{tab:model_comparison}).
This is primarily because \proj uses a single image as the model input, which significantly reduces the compute overhead and, in turn, allows us to use larger models.

\mode{SVC2D} is a prior CE-based AR method that learns the CE pattern jointly with SVC in the downstream model~\cite{ar_iccp, ar_tpami}.
When we apply its CE pattern to our ViT model, it underperforms our decorrelation-based pattern by 1.35\% in AR accuracy (not shown in the table).
This highlights a key limitation of task-specific, end-to-end learned patterns: they are model-dependent and become suboptimal even for the same task when the model architecture evolves.

\begin{table}[t] %
    \caption{Comparison with Previous Systems. We highlight the Top 3 variants for each metric (red, orange, yellow).}
    \centering
    \renewcommand\arraystretch{1.2}
    \resizebox{0.49 \textwidth}{!}{
        \begin{tabular}{l|c|ccc|c}
        \toprule
        \textbf{Model} & \textbf{Input} & \multicolumn{3}{c|}{\textbf{Accuracy (↑)}} & \textbf{Inference} \\
                       &                & \textbf{UCF-101} & \textbf{SSV2} & \textbf{K400} &  \textbf{/sec (↑)} \\
        \midrule
        \proj-S (ours)         & CE             & \cellcolor{tabsecond} 74.65\%      & \cellcolor{tabsecond} 42.38\%            & \cellcolor{tabsecond} 47.58\%            & \makebox[22pt][r]{ \cellcolor{tabfirst} 2282 } \\
        \proj-B (ours)         & CE             & \cellcolor{tabfirst} 79.14\%      & \cellcolor{tabfirst} 45.21\%           & \cellcolor{tabfirst} 54.11\%            &  \makebox[12pt][r]{\cellcolor{tabthird} 760} \\
        SVC2D~\cite{ar_iccp}        & CE             &  \hspace{0.05em}  41.16\%      &  \hspace{0.05em}  23.05\%       & \hspace{0.01em} 26.09\%            & \makebox[16pt][r]{\cellcolor{tabsecond} 2135} \\
        C3D~\cite{tran2015learning}          & Video          & \hspace{0.05em} 62.70\%      &  \hspace{0.05em} 33.48\%       & \hspace{0.05em} 41.66\%         &  541 \\
        VideoMAEv2-ST~\cite{wang2023videomae}    & Video          & \cellcolor{tabthird} 72.54\%      & \cellcolor{tabthird} 39.84\%       & \cellcolor{tabthird} 41.99\%            & 750 \\
        \bottomrule
        \end{tabular}
        }
    \label{tab:model_comparison}
\end{table}

\subsection{\proj Significantly Reduces Edge Energy}
\label{sec:eval:power}
Our energy evaluation is mainly based on edge-server scenarios where the edge device has to transmit all the data to the cloud/server.
In this case, the edge energy is the sum of the sensing and data transmission energy.
We use CamJ~\cite{ma2023camj}, a sensor energy model calibrated against real silicon to model the sensing energy.
The total sensing energy is 220 pJ per pixel (8 bits), of which 95.6\% is contributed to by the ADC and MIPI energy~\cite{ma2023leca}.
The energy overhead introduced by supporting CE is 9 pJ per pixel with 20 MHz pattern stream clock according to our synthesis results.

The wireless energy depends on the technology used for data transmission.
We model short-range (about 10 meters) transmission using passive WiFi, with a transmission energy of 43.04~pJ per pixel~\cite{kellogg2016passive}.
Long-range (over 100 meters) transmission usually uses backscatter Long-Range (LoRa) technology~\cite{talla2017lora}, whose energy is 7.4~$\mu$ J per pixel.

Under $T=16$, \proj reduces the ADC/MIPI and wireless transmission energy by 16$\times$.
\proj achieves a 7.6$\times$ edge energy saving in the short-range scenario and a 15.4$\times$ saving in the long-range scenario.

We also evaluate a scenario where the edge node has a mobile GPU that can execute the downstream model.
For that we measure the energy consumption of the mobile Volta GPU on a Jetson Xavier SoC~\cite{xaviersoc} with a batch size of 1.
In this case the mobile GPU energy dominates the total energy consumption.
Compared to executing video baselines: \mode{VideoMAEv2-ST} and  \mode{C3D} on the edge server, \mode{\proj-S} achieves $1.4\times$ and $4.5\times$ energy saving respectively.
This is primarily because \proj uses single (coded) images to drive the vision model rather than an entire video.

Finally, we compare with a simple compression baseline that spatially downsamples each frame by 16$\times$ (the same compression rate as \proj) using 4$\times$4 average filtering and then processing the compressed data with \mode{VideoMAEv2-ST}.
The resulting AR accuracy is 9.83\%, 6.24\%, and 16.45\% \textit{lower} than \mode{\proj-B} on UCF-101, SSV2, and K400, respectively.

\subsection{Ablation Study}
\label{sec:eval:ablat}
We perform an ablation study by removing various components from \Sect{sec:enc} and \Sect{sec:codesign}, using \proj-S as a baseline and the SSV2 dataset and AR task for evaluation:
\begin{itemize}
    \item Removing pre-training reduces accuracy by 11.39\%.
    \item Replacing the decorrelated pattern with a random pattern further decreases accuracy by 3.43\%.
    \item Replacing the tile-repetitive CE pattern with a global pattern reduces accuracy by 23.74\%.
\end{itemize}

%% file: related.tex
\section{Related Work}

\paragraph{Compressed Edge Sensing.}
Digital-domain compression~\cite{wallace1991jpeg, cheng2018deep} could achieve high compression rate.
However, even with dedicated hardware, classic digital compression consumes nJ/pixel~\cite{polonelli2020energy}, several orders of magnitude higher than the energy of sensing itself.
This issue is further exacerbated in deep-learning-based compression~\cite{cheng2018deep}.
Moreover, digital compression operates after sensor read-out without sensing energy saving.
Instead, in-sensor compression~\cite{ma2023leca, feng2024blisscam, yoshida2019high, zhang2023improving, nair20243d} performs compression before read-out, saving both sensing and transmission energy.
Our method falls under the category of in-sensor compression.

\paragraph{Sensors with CE Support.} 
Coded Exposure (CE) is a computational imaging technique used for enhancing imaging quality (e.g. High Dynamic Range (HDR)~\cite{gulve202339, zhang2020closed} and high-speed imaging~\cite{yoshida2019high}).
We propose a novel use of CE --- for in-sensor compression.
Algorithmically compared to prior CE-based methods~\cite{ar_iccp, ar_tpami, reddy2011p2c2}, we are the first to propose using decorrelation to learn the CE patterns, and have shown that such a pattern out-perform prior methods across multiple tasks.

\paragraph{In-Sensor Computation.}
Vision sensors increasingly integrate computational capabilities, shifting computations to the analog domain. 
Examples include selective reading~\cite{zhang2023improving, feng2024blisscam}, feature extraction~\cite{ma2022hogeye}, computing temporal derivatives~\cite{jee2023time}, and even lightweight neural networks~\cite{likamwa2016redeye, nair20243d, ma2023leca}. 
\proj supports in-sensor CE for compression, and introduces only minor augmentations to the hardware design.

%% file: conclusion.tex
\section{Conclusion}

In-sensor compression through CE, which turns a sequence of video frames into a single coded image, leads to an order of magnitude energy saving on edge sensing while maintaining a task quality comparable with existing video-based methods.
Task-agnostic training of CE patterns through decorrelation can generalize to downstream tasks of different nature, out-performing other CE patterns.